\pdfoutput=1
\documentclass[11pt]{article}

\usepackage[final]{acl} 

\usepackage{times}
\usepackage{latexsym}

\usepackage[T1]{fontenc}

\usepackage[utf8]{inputenc}

\usepackage{microtype}

\usepackage{inconsolata}

\usepackage{graphicx}
\usepackage{amsmath}
\usepackage{algorithm}
\usepackage{algpseudocode}
\usepackage{booktabs}
\usepackage{times}
\usepackage{enumitem}
\usepackage{multirow}
\usepackage{soul}
\usepackage{pgfplots}
\usepackage{hyperref}
\usepackage{url}
\usepackage{subcaption}
\usepackage[most]{tcolorbox}
\usepackage{multicol}

\usepackage{physics}
\usepackage{verbatim}
\usepackage{amssymb}
\usepackage{pifont}
\usepackage{ulem}
\usepackage{mathrsfs}
\usepackage{xspace}
\usepackage{array}
\usepackage{arydshln}
\usepackage{float}
\usepackage{listings}

\useunder{\uline}{\ul}{}
\usepackage{colortbl}
\PassOptionsToPackage{prologue,dvipsnames}{xcolor}
\usepackage[dvipsnames]{xcolor}
\usepackage{pgfmath}
\usepackage{xurl} 
\newcolumntype{L}[1]{>{\raggedright\arraybackslash}p{#1}}

\definecolor{codegreen}{rgb}{0,0.6,0}
\definecolor{codegray}{rgb}{0.5,0.5,0.5}
\definecolor{codepurple}{rgb}{0.58,0,0.82}
\definecolor{backcolour}{rgb}{0.95,0.95,0.92}

\lstdefinestyle{mystyle}{
  commentstyle=\color{codegreen},
  keywordstyle=\color{magenta},
  numberstyle=\tiny\color{codegray},
  stringstyle=\color{codepurple},
  basicstyle=\ttfamily\footnotesize,
  breakatwhitespace=false,
  breaklines=true,
  captionpos=b,
  keepspaces=false,
  showspaces=false,
  showstringspaces=false,
  showtabs=false,
  tabsize=2
}
\lstdefinestyle{promptstyle}{
    basicstyle=\footnotesize\ttfamily,
    breaklines=true,
    breakatwhitespace=false,
    columns=fullflexible,
    keepspaces=true,
    frame=single,
    numbers=none,
    showstringspaces=false,
    literate={`}{\textasciigrave}1
}
\lstdefinestyle{skillstyle}{
  backgroundcolor=\color{lightgray},
  basicstyle=\ttfamily\small,
  language=Python,
  numbers=left,
  numberstyle=\tiny,
  stepnumber=1,
  numbersep=5pt,
  frame=single,
  breaklines=true,
  tabsize=4,
  showstringspaces=false
}

\usepackage{microtype}
\usepackage{cleveref}

\newcommand{\gpt}{\textsc{GPT-4.1}\xspace}
\newcommand{\qwen}{\textsc{Qwen3-4B}\xspace}
\DeclareRobustCommand{\skillany}{{\fontfamily{ppl}\selectfont SkillAny}\xspace}

\title{\skillany: Online Skill Learning for Language Agents \\ at Any Intermediate State}

\author{
Jiaxi Li$^1$, Ke Deng$^1$, Yun Wang$^1$, Jingyuan Huang$^1$, Yucheng Shi$^2$\\
\textbf{Yuanfu Sun$^3$, Qiaoyu Tan$^3$, Jin Lu$^{1}$\textsuperscript{\textdagger}, Ninghao Liu$^{4}$\textsuperscript{\textdagger}}\\
$^1$University of Georgia \quad 
$^2$Tencent America \\
$^3$New York University \quad
$^4$The Hong Kong Polytechnic University
}

\pgfplotsset{compat=1.18}

\begin{document}

\maketitle

\begingroup
\renewcommand{\thefootnote}{\textdagger}
\footnotetext{Co-corresponding authors.}
\endgroup

\begin{abstract}
Language agents increasingly rely on reusable skills
to improve multi-step web automation across related tasks.
A growing line of work studies \textit{online skill learning}, where agents continually induce skills from previous task trajectories and reuse them in future tasks 
on the fly.
However, existing methods mainly reuse skills at the task-level: a fixed set of skills is retrieved based on the initial task instruction and then held fixed throughout execution.
This static strategy is misaligned with web execution, where the appropriate next action depends not only on the task goal but also on the current webpage state, which often transitions into situations that the initial skills fail to cover. 
To address this gap, we introduce \textbf{\skillany}, an online skill learning framework that enables language agents to learn, retrieve, and invoke reusable skills at \textit{any intermediate state} during task execution. Three components jointly enable this capability: sliding-window extraction turns completed trajectories into sub-procedures applicable beyond task boundaries, a dual text--code representation makes each skill both retrievable and executable, and step-level task--state retrieval selects skills that match the goal and the current webpage state as execution unfolds.
Experiments on WebArena across five domains show that \skillany consistently outperforms strong baselines, achieving average success rates of 37.5\% with \gpt and 24.3\% with \qwen, corresponding to relative gains of 10.6\% and 10.0\% over the strongest baseline, respectively.
The code is available at \url{https://github.com/plusnli/online-skill-learning}.
\end{abstract}

\section{Introduction}
\label{sec:intro}

\begin{figure}[t]
    \centering
    \includegraphics[width=\linewidth]{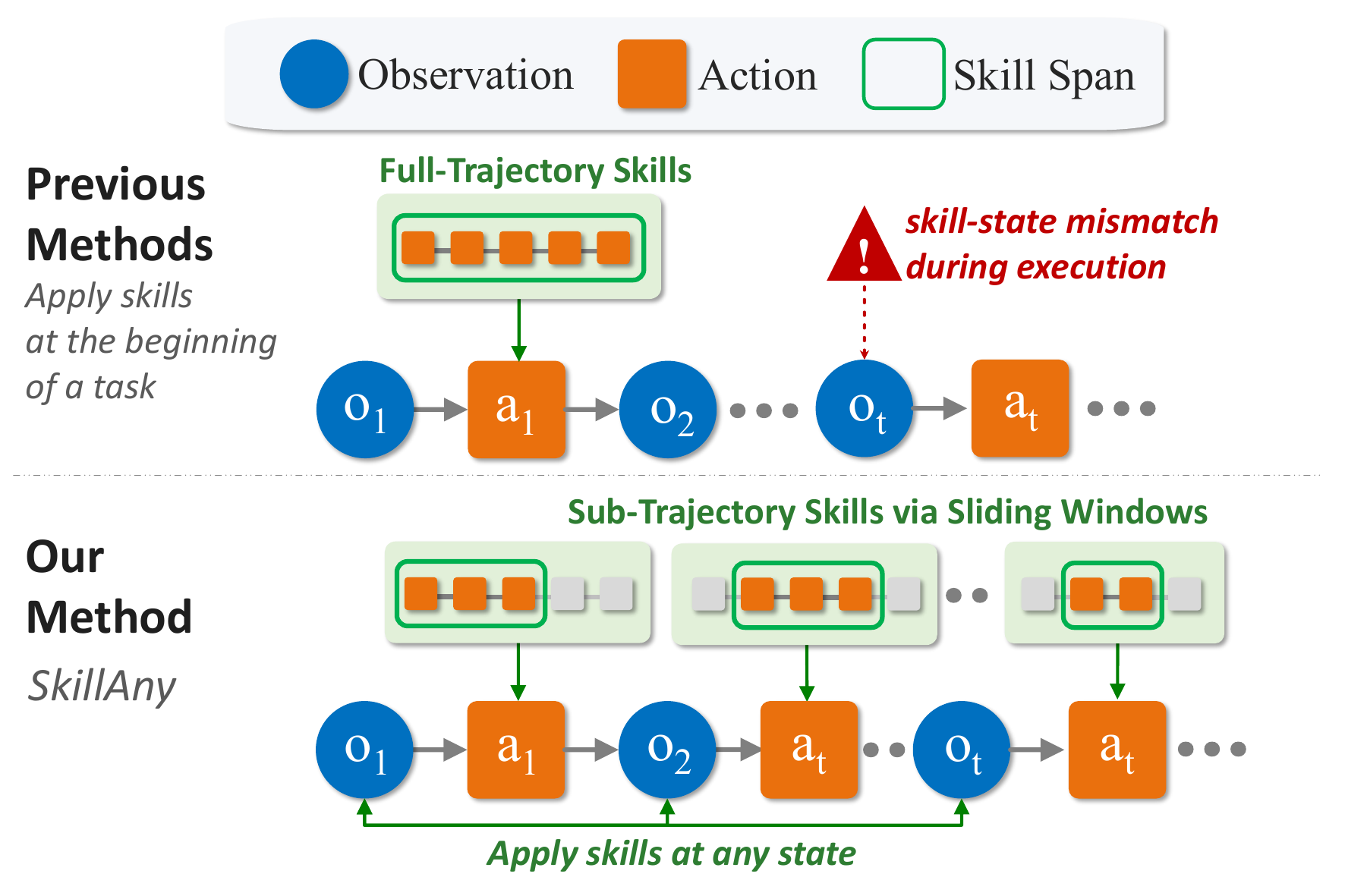}
    \caption{Comparison between previous skill methods and \skillany within the setting of online skill learning.}
    \vspace{-1.5em}
    \label{fig:prev_ours_intro}
\end{figure}

Language agents~\cite{yao2023react, sumers2024cognitive} are increasingly used to solve multi-step web tasks such as information seeking, form filling, and forum interaction on realistic websites~\citep{chae2025web, gu2025is, ning2025survey}.
Although these tasks vary in goals, they often share recurring procedural patterns, such as navigating menus, filling forms, applying filters, and submitting changes.
This observation has motivated a growing line of work on \textit{skill learning} for language agents, where reusable procedural knowledge is summarized and reused in related tasks~\citep{liu2025contextual, zheng2025skillweaver}. By accumulating such skills, agents can amortize repeated procedural discovery and improve across related tasks without relying solely on zero-shot planning~\cite{tack2024online}.

Within this direction, \textit{online skill learning} provides a particularly realistic setting for web agents. Instead of assuming a fixed skill library constructed offline, online methods allow agents to continually induce skills from completed executions and update their skill library as tasks arrive sequentially~\citep{wang2025agent, wang2025inducing, liu2025contextual}.
Compared to relying on a pre-built library constructed offline, this online paradigm more closely matches real-world deployment, where tasks arrive sequentially and agents must improve as they go.

Despite this progress, existing online skill learning methods largely treat skill reuse as a task-level one-shot operation~\citep{wang2025agent, wang2025inducing, liu2025contextual}. Skills are retrieved or injected once based on the initial task instruction and then kept fixed throughout execution. This design is natural if a web task is viewed as a static instruction, but it is insufficient for interactive web automation. 
In web execution, the usefulness of a skill depends not only on the task goal but also on the current webpage state. Consequently, a skill that is useful at the beginning of the task may become irrelevant later, while another skill that was not initially selected may become useful after the agent reaches a new page, form, or interaction context. 
The core limitation is therefore that skill learning and reuse are organized around task boundaries rather than the intermediate execution states where skills actually become applicable.
This raises a central question: \textit{\textbf{how can an online agent learn and reuse the right skill at any intermediate state during execution?}}

Supporting skills at arbitrary intermediate states is non-trivial, since successful reuse depends not only on selecting a skill for the current state but also on whether the library contains skills at an appropriate granularity.
On the one hand, if skills are all at full-trajectory level, they may be occupied by the context from their original tasks but fail to generalize to arbitrary intermediate state~\cite{wang2025agent}.
On the other hand, if skills are at basic single-action level, then they are broadly applicable but too primitive to provide meaningful procedural abstraction. 
This creates a granularity challenge: reusing skills at intermediate state requires skills are compact enough to apply across diverse states, yet structured enough to execute useful consecutive operations. Without skills at this granularity, selecting skills during execution would yield either overly broad workflows that mismatch the current state or low-level actions that offer little benefit over primitive actions.

To address these limitations, we introduce \textbf{\skillany}, an online skill learning framework for learning and invoking reusable skills at \textbf{\textit{any intermediate state}} during task execution, as illustrated in Figure~\ref{fig:prev_ours_intro}. Its name emphasizes that learned skills are available throughout execution, rather than only at task boundaries.
\skillany replaces task-level one-shot reuse with intermediate-state skill learning: it constructs locally applicable skills and selects them anew as the execution state evolves.
After completing a task, \skillany extracts reusable sub-procedures from the trajectory through sliding-window extraction, producing skills at an intermediate granularity.
Each skill is represented as a text–code pair: a natural-language description supports retrieval, while executable code provides support for action. 
When solving a new task, \skillany retrieves step-specific skills conditioned on both the task instruction and the current webpage state, enabling skill support to adapt as execution unfolds.
Together, these designs make skill learning adaptive throughout execution, allowing skills to be learned for and reused at any intermediate state. Our major contributions are summarized below.
\begin{itemize}[itemsep=0pt,topsep=0pt,leftmargin=*]
    \item We study online skill learning for agents under a sequential task-stream setting, where agents can only reuse skills induced from past task trajectories and update the skill library on the fly. 
    \item We introduce \skillany, which moves skill learning beyond task-level static reuse by making learned skills available at any intermediate state during execution.
    \item We realize intermediate-state skill reuse through sliding-window extraction, a dual text--code representation, and step-level retrieval, producing local sub-procedures that are executable and can be selected when they become applicable.
    \item We evaluate \skillany on WebArena across five website domains with two backbone models, showing consistent overall improvements over strong online skill learning baselines in both task success rates and efficiency.
\end{itemize}

\section{Related Work}
\label{sec:related_work}

\subsection{Web Agents and Benchmarks}
Early web agent research~\citep{zheran2018reinforcement, nakano2021webgpt, yao2022webshop} studied how language models interact with browsers to retrieve information and complete tasks in simulated environments. Recent work has scaled web agents toward more realistic settings along several axes: generalist navigation on real-world websites~\cite{deng2023mind2web, he2024webvoyager, pmlr-v235-zheng24e, lai2024autowebglm, hu2025automated, yu2026browseragent}, robustness through memory, workflow induction, and reusable skills~\citep{zheng2024synapse, pmlr-v235-wang24az, wang2025agent, wang2025inducing, zheng2025skillweaver}, 
and benchmarks that evaluate agents under increasingly realistic conditions including visually grounded and conversational navigation~\citep{zhou2024webarena, koh2024visualwebarena, pmlr-v235-lu24e, pmlr-v235-drouin24a, xue2025an, tian2025mmina, yang2026tron, yang2026self, gou2026mindweb}. 
Together, these efforts move web agent research from controlled browser interaction toward dynamic, long-horizon web automation.

\newpage

\subsection{Skill Discovery and Learning}
Recent work explores how language agents can self-improve by discovering and accumulating reusable skills from past executions~\citep{qian2024investigate, yu2025polyskill, ouyang2026skillos, ouyang2026reasoningbank, wang2026skillorchestra, yang2025automated, lu2026contractskill, fang2025memp}. 
Early approaches store procedural knowledge in natural language and adapt it non-parametrically, such as verbal reflections~\citep{shinn2023reflexion} or distilled experiential insights~\citep{zhao2024expel}. 
More recent work formulates reusable skills as workflows~\citep{wang2025agent}, executable programs~\citep{wang2025inducing}, or retrievable past experiences~\citep{liu2025contextual}, with further studies exploring diverse approaches for skill extraction~\cite{zheng2025skillweaver, wei-etal-2026-evo, zhou2025proposeragentevaluator}, organization~\citep{wang2026skillx, li2025mits, xiong-etal-2026-skillpyramid},
and reuse~\citep{wang2026webxskill, jiang2026xskill, wang-etal-2026-reinforcement, qi2026co, ni-etal-2026-trace2skill, liu-etal-2026-harnessing, su-etal-2026-skill}. 
Our work complements these efforts by moving beyond task-level skill reuse, enabling agents to learn and invoke reusable skills at \textit{any intermediate state} during execution.

\section{Preliminaries}
\label{sec:prelim}

\subsection{Task and Skill Formalization}
\label{sec:prelim_formalization}

We consider a sequence of web agent tasks $\mathcal{G} = \{g_i\}_{i=1}^{N}$, where each $g_i$ denotes the natural language instruction specifying the task goal, with a total of $N$ tasks.
When solving the $i$-th task $g_i$, the agent interacts with a web environment over multiple steps, receiving the current webpage observation and executing an action, producing a trajectory $\mathcal{T}_i$, which is an observation-action interleaving sequence of length $H_i$.

The agent maintains a skill library throughout the task sequence. We denote the skill library by $\mathcal{S}_i$ after processing the first $i$ tasks, with $\mathcal{S}_0$ being the initial empty library. Each skill $s \in \mathcal{S}_i$ represents reusable procedural memory induced from previous task executions. After executing task $g_i$, the agent may induce a set of new skills $\Delta \mathcal{S}_i$ from its trajectory and update the library as
\(\mathcal{S}_i = \mathcal{S}_{i-1} \cup \Delta \mathcal{S}_i\).

For evaluation, we use $y_i \in \{0,1\}$ to denote the ground-truth task success signal used for external benchmarking, where $y_i=1$ indicates that the task is correctly solved and $y_i=0$ indicates that the task is not correctly solved.

\begin{figure}[t]
    \centering
    \vspace{-1.1em}
    \includegraphics[width=0.93\linewidth]{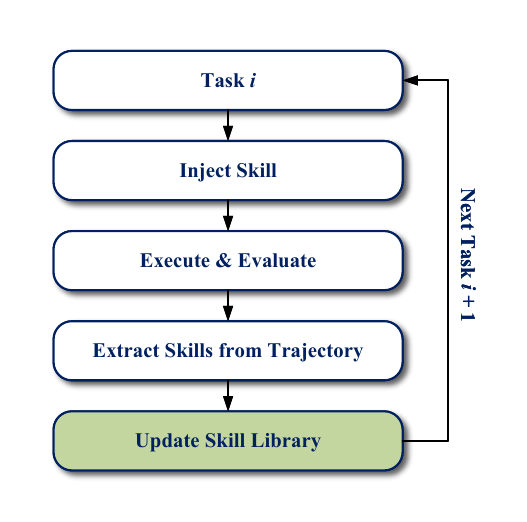}
    \vspace{-1em}
    \caption{The online skill learning setting. The agent solves tasks sequentially, updates the skill library from evaluator-assessed trajectories, and reuses accumulated skills for future tasks.}
    \vspace{-1em}
    \label{fig:online_skill_learning_loop}
\end{figure}

\subsection{Online Skill Learning}
\label{sec:prelim_setup}

Online learning is a sequential learning paradigm in which a learner makes decisions over a stream of examples and uses information revealed from previous rounds to improve future decisions~\citep{cesa2006prediction, shalev2025online}.
In this work, we formulate \textit{online skill learning} for language agents as a task-stream setting in which an agent solves tasks sequentially, updates its skill library from completed trajectories, and reuses only skills induced from past tasks when solving future tasks.
This contrasts with offline skill learning, where a fixed skill library is pre-constructed from a separate set of tasks before being used to assist the agent on held-out evaluation tasks.

\begin{figure*}[t]
    \centering
    \vspace{-1em}
    \includegraphics[width=0.95\textwidth]{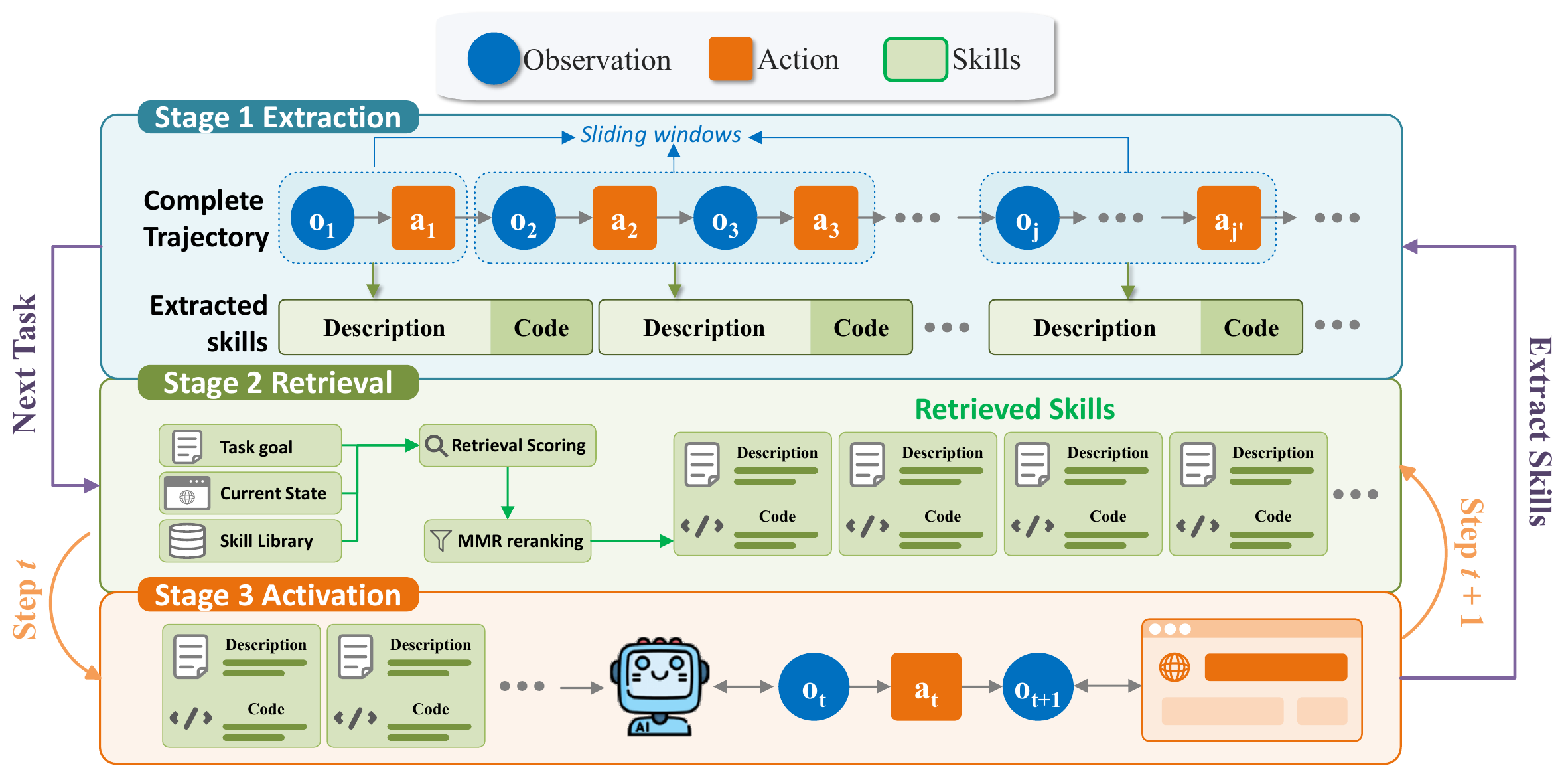}
    \vspace{-0.5em}
    \caption{Overview of \skillany. Completed trajectories are segmented with sliding windows to induce reusable text--code skills applicable from intermediate states. During future tasks, \skillany retrieves skills and checks their applicability at every decision step using the task goal and current webpage state.}
    \vspace{-1em}
    \label{fig:method_pipeline}
\end{figure*}

Figure~\ref{fig:online_skill_learning_loop} depicts the overall setup. 
In line with prior work
~\cite{wang2025inducing, wang2025agent, liu2025contextual}, tasks arrive sequentially: when solving task $g_i$, the agent can only access the skill library accumulated from previous tasks, namely $\mathcal{S}_{i-1}$.
The ground-truth environment signal $y_i$ for the current task is unavailable during both execution and library update, and therefore cannot be used for skill induction or action selection.
The agent must complete the task using only the current instruction, the evolving webpage observations, and skills induced from past tasks.
To support skill induction without access to $y_i$, an evaluator model $E$ is used to assess the completed trajectory after execution: \(\hat{y}_i = E(g_i, \mathcal{T}_i)\), where $\hat{y}_i \in \{0,1\}$ denotes the evaluator's binary correctness judgment for task $g_i$, with $\hat{y}_i=1$ indicating that $E$ judges the completed trajectory to have correctly solved the task, and $\hat{y}_i=0$ indicating that $E$ judges it to have failed.

After executing $g_i$, the agent updates the skill library without observing the ground-truth signal $y_i$. The update can only rely on the task instruction $g_i$, the collected trajectory $\mathcal{T}_i$, and the evaluator-produced proxy judgment $\hat{y}_i$. We formalize skill induction as an update function $U$:
\(\Delta \mathcal{S}_i = U(g_i, \mathcal{T}_i, \hat{y}_i)\), then \(\mathcal{S}_i = \mathcal{S}_{i-1} \cup \Delta \mathcal{S}_i\).
The newly induced skills become available only for subsequent tasks $g_{i+1}, \ldots, g_N$.

The goal of online skill learning is to design an online agent that maximizes cumulative task success rates $y_i$ over the task stream: \(\max_{\pi} \sum_{i=1}^{N} y_i\), where $\pi$ denotes the overall online skill learning agent, including its action policy, skill induction, and skill reuse rules.

\section{Proposed Method}
\label{sec:method}
Building on the online setting in \Cref{sec:prelim_setup}, \skillany addresses two challenges in making learned skills available at any intermediate state: constructing skills that capture local but meaningful procedures, and selecting applicable skills as the webpage state evolves.
To address these challenges, \skillany combines sliding-window extraction and a text--code skill representation with step-level task--state retrieval and reranking.
\Cref{fig:method_pipeline} illustrates the overall pipeline.

Unless otherwise specified, we take task $g_i$ in the online task stream to describe the pipeline of \skillany, and omit the task index $i$ for readability.
Thus, we write the current task as $g$, its trajectory as $\mathcal{T}$, and the current skill library as $\mathcal{S}=\mathcal{S}_{i-1}$.

\subsection{Skill Extraction and Representation}
\label{sec:method_skill_extraction}
We first describe the unit of reuse maintained by \skillany.
Before solving the current task $g$, the agent has access only to the skill library accumulated from previous tasks, denoted as $\mathcal{S}=\{s_k\}_{k=1}^{n}$.
Each skill $s_k$ stores a reusable web procedure and is represented as a text--code pair $s_k=\big(d_k,c_k\big)$, where $d_k$ is a natural-language description used for retrieval and $c_k$ is an executable code function used for action execution.
This text--code representation ties retrieval and execution together: the description abstracts the skill's intent and applicable state, while the code implements the corresponding web operations once the skill is selected. For example, a description such as ``navigate to the account address settings page'' can be paired with code that opens the account menu, clicks the address settings entry, and waits for the target form to load.

After task $g$ is finished, the evaluator produces a binary judgment $\hat{y}$ for its completed trajectory.
We perform skill extraction only when $\hat{y}=1$, i.e., when the evaluator $E$ judges the trajectory to have successfully solved the task.
For such successful trajectories, we revisit the full trajectory $\mathcal{T} = (o_{1}, a_{1}, o_{2}, a_{2}, \ldots, a_{H}, o_{H+1})$, where $H$ is the interaction horizon.
At any step $t \in \{1,\ldots,H\}$, $o_{t}$ represents the current webpage observation that the agent receives, and $a_{t}$ denotes the executed action, forming an observation-action interleaving trajectory.
In web environments, $o_{t}$ can be represented by the textual form of the webpage accessibility tree, which contains structured information about visible elements, their attributes, and possible interaction targets.
The set of primitive actions is provided in Appendix~\ref{app:action_space}.

Rather than storing the entire trajectory as a single task-level skill, we decompose it into local segments that can be reused from intermediate states in future tasks.
We then apply a set of sliding windows over the trajectory to obtain candidate segments.
For each window length $l \in \mathcal{L}$, we enumerate candidate segments \(w_{t,l} = (o_{t}, a_{t}, \ldots, a_{t+l-1}, o_{t+l})\), where $t$ denotes the window's starting timestep.

Sliding windows enable the extraction of reusable skills at an intermediate granularity. Full trajectories often encode an entire task and are too specific to be reused at a later intermediate state, while individual actions are too fine-grained to capture meaningful procedures. Windowed segments instead correspond to local but reusable subroutines, such as opening a settings page, filling a short form, or applying a filter.

Multiple candidate segments $w_{t,l}$ are batched into an LLM call. The LLM independently judges whether each segment captures a reusable state-contingent procedure and, if so, converts it into a skill $s_k = \big(d_k, c_k\big)$.
Following ASI~\cite{wang2025inducing}, we verify each induced skill by replacing its corresponding primitive action segment in the original trajectory with the skill and executing the rewritten trajectory in the environment.
Only the skills whose substituted trajectories are still judged successful by the evaluator are added to the library.
Together, this sliding-window extraction and verification process yields skills that are compact enough to be invoked from intermediate execution states, while remaining executable and semantically meaningful.
Once added to the library, these verified text--code skills become candidates for step-level retrieval in subsequent tasks.

\subsection{Skill Retrieval at Intermediate States}
\label{sec:method_dyretr}
Given the verified skill library, \skillany selects skills anew at every intermediate state, rather than choosing a fixed set only once at the beginning of a task.
At execution step $t$ of task $g$, the agent observes the current web state $o_{t}$. 
As raw web states such as accessibility trees can be verbose, we first truncate the accessibility-tree input and use an LLM to obtain a compact state summary $r_{t}=\mathrm{Summarize}(o_{t})$.
The resulting summary serves as the state-side retrieval query, while the original task instruction $g$ provides the goal-side query.

\paragraph{Relevance Retrieval.}
To retrieve appropriate skills at step $t$, we do relevance retrieval over the skill library $\mathcal{S}$.
For each skill $s_k = (d_k, c_k)$, we compute a relevance score combined task-state:
\begin{equation}
    \label{eq:task_state_score}
    \begin{aligned}
        \operatorname{score}_{t}(s_k) ={}& \alpha \, \cos\big(\phi(g), \phi(d_k)\big) \\
        &+ (1-\alpha) \, \cos\big(\phi(r_{t}), \phi(d_k)\big).
    \end{aligned}
\end{equation}
Here $\phi(\cdot)$ maps text into the embedding space, and $\cos(\mathbf{u}, \mathbf{v}) = \mathbf{u}^{\top}\mathbf{v} / (\|\mathbf{u}\|\|\mathbf{v}\|)$ denotes cosine similarity between two embeddings $\mathbf{u}$ and $\mathbf{v}$.
The coefficient $\alpha$ is balances the overall task instruction and the current state. The first term measures alignment with the task goal, while the second term measures applicability to the current state. We first keep the top-$M$ candidate skills according to their relevance score $\operatorname{score}_{t}(s_k)$,
and then pass them to the reranking stage described below.
This stage filters the library to skills that are broadly relevant to the current task and state. 

\begin{table*}[t]
    \centering

    \caption{Main success rates (\%) on WebArena. We list the task success rates for five website domains of WebArena. \# Steps denotes the average number of steps to complete each task, and SR denotes the averaged success rate overall. The best result is shown in \textbf{bold}, and the second-best result is \underline{underlined}.}

    \fontsize{9.3pt}{10.8pt}\selectfont

    \scalebox{1.1}{%
        \begin{tabular}{ll|cc|ccccc}
            \toprule
            Model & Method & \# Steps & SR & Shopping & Admin & Reddit & Gitlab & Map \\
            \midrule

            \multirow{5}{*}{\gpt}
            & Vanilla & 6.0 & 28.1 & 28.9 & 35.7 & 21.7 & 27.8 & 21.1 \\
            & AWM     & 5.9 & 27.7 & 26.7 & 36.3 & 24.5 & 28.3 & 17.4 \\
            & ASI     & 5.2 & 32.9 & 29.4 & \underline{41.2} & \underline{34.0} & 29.4 & \underline{29.4} \\
            & CER     & 6.4 & \underline{33.9} & \underline{31.0} & 38.5 & 31.1 & \textbf{37.2} & 28.4 \\
            & \skillany (Ours) & 4.8 & \textbf{37.6} & \textbf{34.8} & \textbf{47.8} & \textbf{35.8} & \underline{34.4} & \textbf{32.1} \\

            \midrule

            \multirow{5}{*}{\qwen}
            & Vanilla & 6.3 & 16.5 & 21.9 & 14.8 & 12.3 & 16.7 & 13.8 \\
            & AWM     & 5.7 & 15.6 & 19.8 & 12.6 & 13.2 & 18.3 & 11.0 \\
            & ASI     & 5.9 & 20.7 & 22.5 & \underline{22.5} & 18.9 & 22.2 & 13.8 \\
            & CER     & 6.5 & \underline{22.3} & \underline{23.5} & 20.9 & \underline{19.8} & \textbf{27.8} & \underline{15.6} \\
            & \skillany (Ours) & 5.6 & \textbf{24.2} & \textbf{25.1} & \textbf{24.7} & \textbf{22.6} & \underline{26.7} & \textbf{19.3} \\

            \bottomrule
        \end{tabular}%
    }
    \vspace{-0.5em}
    \label{tab:main_results}
\end{table*}

\paragraph{MMR Reranking.}
The relevance retrieval stage produces a top-$M$ candidate set whose members are individually relevant to the current task and state. 
However, because skills are extracted from overlapping sliding windows, many high-scoring candidates may correspond to near-duplicate local procedures with slightly different boundaries or contexts. 
Directly passing the top-ranked skills to the agent can therefore allocate multiple skill slots to the same procedural pattern, leaving fewer distinct options for the next decision. 
To avoid this redundancy while preserving relevance, we apply Maximal Marginal Relevance (MMR)~\cite{carbonell1998use} within the relevance-filtered candidate set. 
This reranking step is not a replacement for relevance retrieval: the relevance score keeps each selected skill grounded in the current state, while the diversity penalty discourages selecting skills that overlap with those already chosen.
Starting from an empty set $\mathcal{A}_{t}$, we greedily add skills until $|\mathcal{A}_{t}|=5$, where each next skill \(s_k\) is selected according to
\begin{equation}
    \label{eq:mmr}
    \begin{aligned}
        \operatorname{MMR}_{t}(s_k) ={}&
        \lambda \, \operatorname{score}_{t}(s_k) \\
        &\hspace{-2em}-(1-\lambda)
        \max_{s_{k'} \in \mathcal{A}_{t}}
        \operatorname{sim}(d_k, d_{k'}).
    \end{aligned}
\end{equation}
Here $\operatorname{sim}(d_k, d_{k'}) = \cos(\phi(d_k), \phi(d_{k'}))$ denotes the cosine similarity between the two skill descriptions in embedding space and serves as a proxy for procedural overlap. 
$\lambda$ balances relevance
and coverage among selected skills.
The resulting set $\mathcal{A}_{t}$ is the step-specific skill set activated for the agent's current decision.

\subsection{Skill Injection and Execution}
\label{sec:method_exec}
After retrieval and reranking, the selected skill set $\mathcal{A}_{t}$ is exposed to the agent only for the current decision step $t$. 
For each retrieved skill, we provide its description $d_k$ and callable code $c_k$ as additional action support.
Then the agent checks the applicability of the skills, and chooses either a skill or a primitive action at the current step.
This step-level injection is what makes skills available at any intermediate state: the support presented to the agent changes with the evolving webpage, without exposing the full skill library at every decision step.
After the task is completed, the collected trajectory is evaluated and processed by the extraction procedure illustrated in \Cref{sec:method_skill_extraction}. 
The resulting verified skills are added to the corresponding domain-specific library and become available for subsequent tasks, starting from $g_{i+1}$.

\begin{figure*}[!t]
    \centering
    \includegraphics[width=1\textwidth]{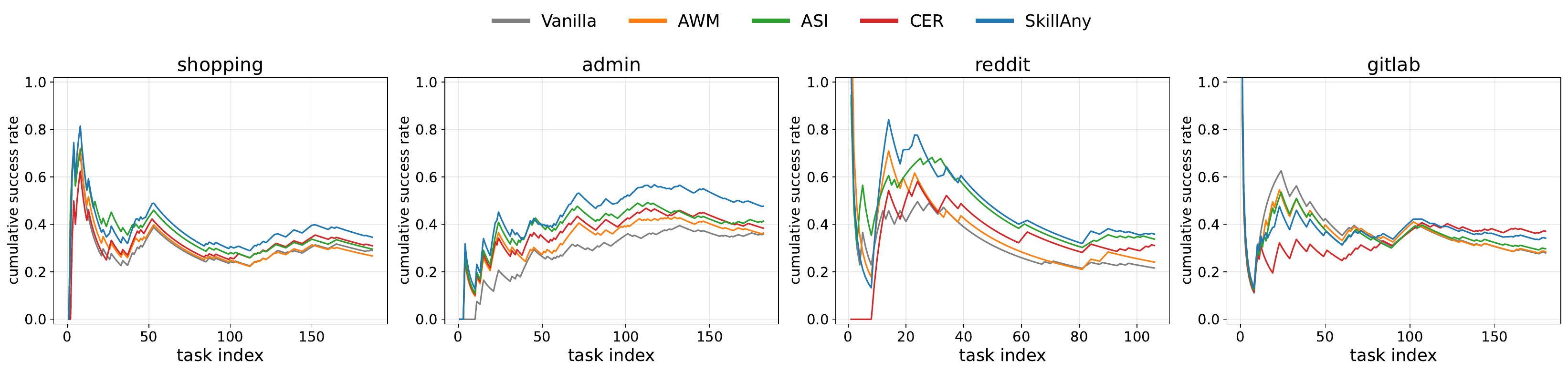}
    \caption{Cumulative success rates over the online task stream with backbone model \gpt on four WebArena domains. The x-axis denotes the remapped within-domain task index and sorting by the original WebArena task IDs. \skillany generally maintains stronger cumulative performance as more tasks are processed, showing the benefit of making learned skills available throughout execution.}
    \label{fig:analysis_cumu_sr_gpt41}
    \vspace{-1em}
\end{figure*}

\section{Experiments}
\label{sec:exp}

\subsection{Experiment Setup}
\label{sec:exp_setup}

\paragraph{Benchmark.}
We evaluate on WebArena~\citep{zhou2024webarena}, a representative and realistic web agent benchmark whose structure is well suited to our online skill learning setting.

WebArena spans five website domains, Shopping, Admin, Reddit, Gitlab, and Map, where tasks within each domain typically share similar website interface and interaction conventions.
This domain structure naturally supports our domain-wise continual skill acquisition: for a given website domain, after completing a task, the agent extracts skills from the resulting trajectory and reuses them for subsequent tasks in the same domain.
Since a small number of WebArena tasks require interactions across multiple websites, we exclude such tasks and focus on single-domain tasks.
Accordingly, we maintain a separate skill library for each domain to avoid cross-domain interference. We list the detailed task indices within each website domain in Appendix~\ref{app:task_indices}.
The evaluation by WebArena environment is based on a binary success reward: the reward is $1$ if the task is correctly solved, and $0$ otherwise.

\vspace{-0.7em}
\paragraph{Baseline Methods.}
We compare \skillany with four baselines.
\textit{Vanilla} is a skill-free baseline that solves each task independently without maintaining or reusing skills across tasks.
We then compare with three baseline methods within the paradigm of online skill learning: Agent Workflow Memory (AWM)~\cite{wang2025agent}, Agent Skill Induction (ASI)~\cite{wang2025inducing}, and Contextual Experience Replay (CER)~\cite{liu2025contextual}.
These methods can accumulate reusable memory from past trajectories and apply it to future tasks.
In our comparison, they primarily instantiate task-level static reuse: relevant workflows, programmatic skills, or past experiences are selected based on the task context and then used as fixed support during execution.
Specifically, AWM stores natural-language workflows, ASI induces executable programmatic skills, and CER retrieves relevant past experiences for decision support.
For AWM and CER, we adopt their online variants, ensuring that all skill-based methods accumulate experience from empty over the same task stream.
We do not involve offline skill learning methods such as SkillWeaver~\citep{zheng2025skillweaver}, because they constructs skill libraries through a separate offline exploration phase before evaluation, rather than learning from empty under our online skill protocol.

\paragraph{Implementation details.}
We report results using \gpt~\cite{achiam2023gpt} and \qwen~\cite{yang2025qwen3} as the backbone models.
For both our method \skillany and the baselines, when using either \gpt or \qwen as the backbone LLM, we use the same model for all LLM-based components, including skill induction, trajectory summarization, execution planning, and evaluation.
For CER, we implement the experience buffer, experience synthesis, and retrieval modules following the original paper~\cite{liu2025contextual}.
For \skillany, we segment the resulting trajectory with sliding windows of length $\mathcal{L} \in \{2,3,4,5\}$ for skill extraction.
During task execution, skills are retrieved at each decision step using the task--state relevance score defined in \Cref{eq:task_state_score}, with $\alpha$ set to $0.5$, followed by reranking with the MMR objective in \Cref{eq:mmr}, where $\lambda = 0.7$.
Detailed configuration is given in Appendix \ref{app:prompts} and \ref{app:param_config}.

\subsection{Main Results}
\label{sec:main_results}

\Cref{tab:main_results} reports the success rates on WebArena.
Overall, \skillany achieves the best average success rate under both backbones, reaching 37.5\% with \gpt and 24.3\% with \qwen.
Compared with the strongest baseline CER, \skillany improves the overall SR by 3.6 points with \gpt and 2.2 points with \qwen, showing the benefit of learning locally applicable skills and making them available at intermediate states beyond static task-level reuse.

The gains are broadly distributed across domains. 
With \gpt, \skillany achieves the best performance on four of the five domains, including a notable improvement on Admin from 41.4\% to 47.7\%.
A similar trend holds for \qwen, while Gitlab remains the main exception. 
We hypothesize that Gitlab tasks often involve version-control operations with persistent repository preconditions, such as forking and merge-request operations.
Since \skillany learns local rather than whole-task skills, it may be less effective for such tasks than methods that preserve complete task-level procedures.

\paragraph{Execution Efficiency.}
We also examine execution efficiency through average step count. Across both backbone models, \skillany completes tasks with fewer steps than the baselines.
With \gpt, it uses 4.8 steps on average, compared with 6.0 for Vanilla, 5.2 for ASI, and 6.4 for CER.
With \qwen, it reduces the average step count by 11.1\% relative to Vanilla and 13.8\% relative to CER. 
This efficiency gain arises because one skill can execute a short procedure composed of multiple primitive browser actions, such as a sequence of clicks and fills, thereby replacing repeated low-level interactions with a higher-level reusable action.

\subsection{Online Performance Analysis}
A central motivation of \skillany is to improve skill reuse throughout the online task stream.
\Cref{fig:analysis_cumu_sr_gpt41} shows the cumulative success rate with \gpt on four WebArena domains, where tasks are ordered by their original WebArena IDs and reindexed within each domain.
Overall, \skillany generally stays above the baselines, with especially clear advantages on Admin and Reddit.
Although the curves are not monotonic because later tasks may be harder or less aligned with previously accumulated skills, \skillany often remains on the upper envelope, suggesting that intermediate-state skill learning helps the agent exploit its growing library throughout execution.
The smaller margin on Gitlab is consistent with its reliance on persistent repository-specific preconditions, which can limit the transferability of local procedural skills.

\subsection{Ablation Study}
\label{sec:ablation}

We conduct ablation studies with \gpt on the Shopping, Reddit, and Map websites, using the same task sets and task order as the main experiments.
We first examine the interaction between skill granularity and retrieval strategy, then ablate MMR reranking.

\paragraph{Skill Granularity and Retrieval Strategy.}
\label{sec:ablation_factorial}
We conduct a $2\times2$ factorial comparison of full-trajectory versus sliding-window skills under task-level static retrieval and \skillany's step-level task--state retrieval, using the same skill libraries and retrieval budgets within each granularity. Static retrieval selects skills once from the task description at the beginning, without MMR, and injects them only into the initial context, whereas the dynamic condition follows \skillany by performing task--state retrieval and MMR at every step. As shown in \Cref{tab:rebuttal_factorial}, sliding-window skills underperform full-trajectory skills with static retrieval but achieve the best dynamic results, gaining 7.2, 3.4, and 3.8 points on Shopping, Reddit, and Map, compared with only 0.8, 0.6, and 1.1 points for full-trajectory skills. \textbf{This interaction captures the central design principle of \skillany: skills constructed for intermediate states become most useful when they can be selected at the particular states where they apply.}
Appendix~\ref{app:additional_ablations} provides further retrieval-signal and granularity ablations.

\begin{table}[t]
    \centering
    \small
    \setlength{\tabcolsep}{4pt}
    \caption{Success rates (\%) for skill granularity and retrieval strategy with \gpt. Static and dynamic denote the strategies defined in Section~\ref{sec:ablation}. $\Delta$ is the dynamic-minus-static gain in percentage points. The best and second-best success rates within each domain are \textbf{bold} and \underline{underlined}, respectively.}
    \label{tab:rebuttal_factorial}
    \begin{tabular}{@{}llrrr@{}}
        \toprule
        Domain & Skill granularity & Static & Dynamic & $\Delta$ \\
        \midrule
        \multirow{2}{*}{Shopping} & Full trajectory & 30.8 & \underline{31.6} & +0.8 \\
        & Sliding window & 27.4 & \textbf{34.6} & +7.2 \\
        \midrule
        \multirow{2}{*}{Reddit} & Full trajectory & 34.3 & \underline{34.9} & +0.6 \\
        & Sliding window & 32.5 & \textbf{35.9} & +3.4 \\
        \midrule
        \multirow{2}{*}{Map} & Full trajectory & 30.3 & \underline{31.4} & +1.1 \\
        & Sliding window & 28.5 & \textbf{32.3} & +3.8 \\
        \bottomrule
    \end{tabular}
\end{table}

\paragraph{Ablation Study on MMR Reranking.}
\label{sec:ablation_mmr}
We compare relevance-only top-$M$ retrieval with MMR reranking at different $\lambda$ values. As shown in \Cref{tab:ablation_mmr}, every MMR variant improves over relevance-only retrieval, and $\lambda=0.7$ performs best across all three domains, showing the benefit of preserving skill diversity while emphasizing relevance.

\begin{table}[t]
    \centering
    \caption{Ablation study on MMR reranking with model \gpt on websites Shopping, Reddit, and Map.}
    \label{tab:ablation_mmr}
    \resizebox{\columnwidth}{!}{%
    \begin{tabular}{lccc}
        \toprule
        Reranking setting & Shopping & Reddit & Map \\
        \midrule
        w/o MMR, top-$M$ relevance & 31.1 & 31.4 & 30.8 \\
        w/ MMR ($\lambda=0.5$) & \underline{33.9} & \underline{34.8} & \underline{32.1} \\
        w/ MMR ($\lambda=0.7$) & \textbf{34.6} & \textbf{35.9} & \textbf{32.3} \\
        w/ MMR ($\lambda=0.9$) & \underline{33.9} & 34.4 & 31.7 \\
        \bottomrule
    \end{tabular}}
\end{table}

\section{Conclusion}
\label{sec:conclusion}
We present \skillany, an online skill learning framework that enables language agents to learn and reuse skills at any intermediate state rather than only at task boundaries.
By extracting local skills from sliding windows of evaluator-assessed trajectories and selecting them at each step using both task and state information, \skillany provides adaptive support throughout execution.
Results on WebArena show strong performance across five domains with two backbone models, suggesting that intermediate-state skill learning is a practical approach to improving web agents based on both proprietary and open-source models.

\section*{Limitations}
This work still has some limitations. 
First, our experiments are conducted on WebArena, which provides realistic multi-step web tasks but still covers a limited set of website domains, interaction patterns, and agent actions. Evaluating \skillany on broader web environments would further validate its generality.
Second, our study focuses on non-parametric skill accumulation and reuse, without exploring how the learned skills could be integrated with model fine-tuning or long-term agent personalization.
We leave these directions for future work.

\section*{Ethical Considerations}
This work studies online skill learning for language agents in web environments. 
Our experiments are conducted on WebArena and do not involve human subjects, private user data, or interactions with live third-party websites. 
Nevertheless, more capable web agents may raise potential concerns if deployed without appropriate safeguards, since automated agents could perform unintended actions, access sensitive information, or violate website usage policies. 
We therefore view \skillany as a research framework for controlled environments, and practical deployment should include permission checks, action constraints, and monitoring. The learned skills should also be validated before reuse in safety-critical settings.

\bibliography{custom}

\clearpage

\appendix
\section{Experiment Details}
\label{app:exp_details}

\subsection{Agent Action Space}
\label{app:action_space}
\Cref{tab:action-space} shows the default base action space the web navigation agents we employed in all the experiments, within the WebArena environment. 
This action space remains the same for our method and all baseline methods, including vanilla, AWM, ASI, CER, and our method \skillany.

\begin{table}[!h]
    \centering
    \small
    \setlength{\tabcolsep}{3pt}
    \renewcommand{\arraystretch}{1.05}
    \caption{Base primitive action space for web agents throughout our experiments in WebArena.}
    \begin{tabular}{@{}L{0.43\linewidth}L{0.51\linewidth}@{}}
        \toprule
        \textbf{Action Type} & \textbf{Description} \\
        \midrule

        \nolinkurl{noop(wait_ms)} & Do nothing for a specified time. \\
        \nolinkurl{click(elem)} & Click an element. \\
        \nolinkurl{hover(elem)} & Hover over an element. \\
        \nolinkurl{fill(elem, value)} & Type into an element. \\
        \nolinkurl{keyboard_press(key_comb)} & Press a key combination. \\
        \nolinkurl{scroll(x, y)} & Scroll horizontally or vertically. \\
        \nolinkurl{select_option(elem, options)} & Select one or multiple options. \\

        \midrule

        \nolinkurl{goto(url)} & Navigate to a URL. \\
        \nolinkurl{go_back()} & Navigate to the previous page. \\
        \nolinkurl{go_forward()} & Navigate to the next page. \\

        \midrule

        \nolinkurl{new_tab()} & Open a new tab. \\
        \nolinkurl{tab_close()} & Close the current tab. \\
        \nolinkurl{tab_focus(index)} & Bring a tab to the front. \\

        \midrule

        \nolinkurl{send_msg_to_user(text)} & Send a message to the user. \\
        \nolinkurl{report_infeasible(reason)} & Notify the user that the instruction is infeasible. \\
        \bottomrule
    \end{tabular}
    \label{tab:action-space}
\end{table}

\subsection{Task Indices for Website Domains}
\label{app:task_indices}
For reproducibility, we provide the task indices used for each WebArena website domain. 
We remove all cross-site tasks to ensure that skills are extracted and reused within the same website domain, thereby preventing cross-domain skill transfer from confounding the evaluation.
After this filtering, we use 764 single-domain tasks in total: 187 Shopping, 182 Admin, 106 Reddit, 180 GitLab, and 109 Map tasks. 
The detailed task indices for each domain are listed below.

\begin{itemize}[leftmargin=1.2em]
    \item \textbf{Shopping}: 
    21--26, 47--51, 96, 117--118, 124--126, 141--150, 158--167, 188--192, 225--235, 238--242, 260--264, 269--286, 298--302, 313, 319--338, 351--355, 358--362, 368, 376, 384--388, 431--440, 465--469, 506--521, 528--532, 571--575, 585--589, 653--657, 689--693, 792--798.

    \item \textbf{Admin}: 
    0--6, 11--15, 41--43, 62--65, 77--79, 94--95, 107--116, 119--123, 127--131, 157, 183--187, 193--204, 208--217, 243--247, 288--292, 344--348, 374--375, 423, 453--464, 470--474, 486--505, 538--551, 676--680, 694--713, 768--782, 790.

    \item \textbf{Reddit}: 
    27--31, 66--69, 399--410, 580--584, 595--652, 714--735.

    \item \textbf{GitLab}: 
    44--46, 102--106, 132--136, 156, 168--182, 205--207, 258--259, 293--297, 303--312, 314--318, 339--343, 349--350, 357, 389--398, 411--422, 441--452, 475--485, 522--527, 533--537, 567--570, 576--579, 590--594, 658--670, 736, 742--756, 783--789, 799--811.

    \item \textbf{Map}: 
    7--10, 16--20, 32--40, 52--61, 70--76, 80--93, 98--101, 137--140, 151--155, 218--224, 236--237, 248--257, 287, 356, 363--367, 369--373, 377--383, 757--758, 761--767.
\end{itemize}

\subsection{Prompts for LLM-Based Components}
\label{app:prompts}
In this subsection, we list the prompts we give to LLM-based components involved in \Cref{sec:method}.

\subsubsection{Prompts for Trajectory Assessment.}
Here are the prompts we give to the trajectory evaluator model $E$ to assess whether the current trajectory successfully complete the task, as demonstrated in both \Cref{sec:prelim_setup} and \Cref{sec:method_skill_extraction}. They are used not only for our method, but also for other baseline methods AWM, ASI, and CER introduced in \Cref{sec:exp_setup}, as they all require the evaluator model $E$ to judge their trajectories.

\noindent\textbf{System Prompt.} The system prompt requires the evaluator model $E$ to give judgement "success" or "failure" based on the user prompt input.
\begin{lstlisting}[style=promptstyle]
You are an expert in evaluating the performance of a web navigation agent. The agent is designed to help a human user navigate a website to complete a task. Given the user's intent, the agent's action history, the final state of the webpage, and the agent's response to the user, your goal is to decide whether the agent's execution is successful or not.

There are three types of tasks:
1. Information seeking: The user wants to obtain certain information from the webpage, such as the information of a product, reviews, map info, comparison of map routes, etc. The bot's response must contain the information the user wants, or explicitly state that the information is not available. Otherwise, e.g. the bot encounters an exception and respond with the error content, the task is considered a failure. Besides, be careful about the sufficiency of the agent's actions. For example, when asked to list the top-searched items in a shop, the agent should order the items by the number of searches, and then return the top items. If the ordering action is missing, the task is likely to fail.
2. Site navigation: The user wants to navigate to a specific page. Carefully examine the bot's action history and the final state of the webpage to determine whether the bot successfully completes the task. No need to consider the bot's response.
3. Content modification: The user wants to modify the content of a webpage or configuration. Carefully examine the bot's action history and the final state of the webpage to determine whether the bot successfully completes the task. No need to consider the bot's response.

*IMPORTANT*
Format your response into two lines as shown below:

Thoughts: <your thoughts and reasoning process>"
Status: "success" or "failure"
\end{lstlisting}

\noindent\textbf{User Prompt.} Here is the user prompt given to the evaluator model $E$. For the placeholders in this prompt, \texttt{intent} is the task goal, \texttt{last-actions} is the action history of the agent, \texttt{cap} is the final state of the webpage, and \texttt{response} is the response extracted from the last action that the agent gives to the user.
\begin{lstlisting}[style=promptstyle]
User Intent: {intent}

Action History:
{last-actions}

The detailed final state of the webpage:

```md
{cap}
```

Bot response to the user: {response if response else "N/A"}.
\end{lstlisting}

\subsubsection{Prompts for Skill Induction.}
Here we list the prompts use for skill extraction in \Cref{sec:method_skill_extraction}. Given the trajectory windows segmented sliding windows, this skill-induction prompt extracts reusable, single-page-callable sub-routines from successful trajectories and emits each as an executable Python function with a retrieval-friendly description.

\noindent\textbf{System Prompt.}
\begin{lstlisting}[style=promptstyle]
You are a proficient web-automation engineer. You judge whether short slices of a successful web trajectory are reusable sub-routines, and when they are, you emit a small Python function that implements the routine. Follow the user instruction's rules and output format exactly.
\end{lstlisting}

\noindent\textbf{User Prompt.}
\begin{lstlisting}[style=promptstyle]
You will be shown several action windows extracted from a successful web task trajectory by a sliding window of length 2, 3, 4, or 5 steps. Each step is a short thought followed by one or more action calls (e.g. click, fill, select_option).

For each window you must decide:

1. Is the window a *reusable* sub-routine?
   A reusable window:
   - Performs a recognizable web operation that could occur on other tasks (e.g. searching a product, applying a price filter, posting a comment, opening a user profile).
   - Is general enough to apply with different inputs: variable parts (search queries, usernames, element ids that obviously vary across tasks) become function arguments with descriptive names. Windows that depend on one-off element ids or task-specific text that cannot be parameterized are NOT reusable.
   - Contains 2 to 5 action steps.

   Single-page-state callability (IMPORTANT): the agent that will invoke this skill observes only the CURRENT web page at call time. EVERY element ID the skill takes as an argument must be readable from the single accessibility tree visible to the agent at the moment of call.
   - Strongly prefer skills whose argument IDs (button IDs, field IDs, option IDs) are all simultaneously visible on one page state.
   - REJECT skills that require an ID which appears only AFTER a page transition the skill itself triggers. The skill may navigate internally, but the caller must still supply that future ID upfront - and the caller cannot observe pages it has not yet reached. There is NO valid exception.
     * Callable example: "fill title + fill body + click submit" on a single submission form - all three IDs are visible simultaneously on that one page.
     * NOT callable: "click combobox, click option, fill title, fill body, click submit" - the option ID only appears after the combobox is opened, so it is not readable at the moment the routine is called.

2. If reusable, produce:
   - description: a single sentence that MUST contain both
     (a) a precise action verb + object (e.g. "submit a forum post", "apply a price filter", "open a forum-selection combobox", "fill in the title and body"); and
     (b) the typical page context where this routine runs (e.g. "on a forum submission form", "on a product listing page", "in an opened combobox").
     The description embedding is cosine-matched to a page-state summary written in the same operational vocabulary, so generic phrasing like "Performs several clicks" will hurt retrieval.
   - code: a Python function that implements the routine.

Code constraints:
- Use ONLY the following actions: click, fill, hover, keyboard_press, scroll, tab_focus, new_tab, tab_close, go_back, go_forward, goto, send_msg_to_user, report_infeasible, select_option.
- Function arguments must be primitive types (str, int, list of str). No callbacks.
- No try / except.
- Do NOT hardcode user-facing messages inside `send_msg_to_user`; if the routine ends with a message, take it as a `message` argument.

Output format - return a single JSON array, one object per window in the same order they were given. Schema:

[
  {"window_idx": 0, "reusable": true, "func_name": "search_product", "description": "...", "code": "def search_product(query):\n    click('search')\n    fill('search', query)\n    keyboard_press('Enter')\n"},
  {"window_idx": 1, "reusable": false}
]

Only output the JSON array, no surrounding prose, no markdown fences.
\end{lstlisting}

\subsubsection{Prompt for Web Summarization.}
Here is the prompt used for summarizing the webpage state $r_{i,t}=\mathrm{Summarize}(o_{i,t})$ for $i$-th task at execution step $t$, as indicated in \Cref{sec:method_dyretr}. This is the system prompt given to the summarizer. The user input contains a truncated textual accessibility tree of the current webpage, together with its URL and title.
\begin{lstlisting}[style=promptstyle]
You are a state summarizer for a web agent whose action library is indexed by descriptions like 'submit a forum post on a submission form' or 'apply a price filter on a product listing page'. Your summary will be cosine-matched against such skill descriptions, so use the SAME operational vocabulary they do.

Given the current page's accessibility tree (axtree) plus the URL and title, produce ONE short paragraph (1-2 sentences) that:
1. Names the kind of page in operational terms (e.g. 'forum submission form', 'product listing page', 'opened forum-selection combobox', 'post-detail page with comment section').
2. Lists the action verbs this page ENABLES right now - i.e. what sub-routines could plausibly run on this exact state. Use verb + object phrasing (e.g. 'submit a post', 'select a forum', 'fill in the title and body', 'open the sort menu', 'apply a filter').

Do NOT enumerate every visible element, do NOT describe pure visuals (colors, layout), and do NOT mention task instructions or speculate about future steps. Output only the summary text.
\end{lstlisting}

\subsubsection{Prompt for Skill Activation and Execution.}
Here is the user prompt we use to make the web agent make the next-step decision as illustrated in \Cref{sec:method_exec}. 
\begin{lstlisting}[style=promptstyle]
## Retrieved Skills
The following {N} high-level skills were retrieved as candidates for your next sub-routine. If one's intent matches what you need (e.g., walking vs. driving) and the required arguments are visible in the accessibility tree, prefer calling it in a single action. Otherwise proceed with primitive actions - either way, keep making progress toward the goal.

[signature and document description of every retrievd skills.]
\end{lstlisting}

\subsection{Parameter Configuration}
\label{app:param_config}

Table~\ref{tab:param_config} summarizes the main parameter configuration used in \skillany and the experimental setup.
Blank entries indicate parameters that are mentioned in the paper but not explicitly specified.

\begin{table}[!h]
    \centering
    \small
    \caption{Parameter configuration of \skillany and the experimental setup. Blank entries indicate parameters that are mentioned in the method but not explicitly specified in the current paper.}
    \resizebox{\columnwidth}{!}{
    \begin{tabular}{ll}
        \toprule
        Parameter & Value \\
        \midrule
        Sliding-window lengths $\mathcal{L}$ & $\{2,3,4,5\}$ \\
        Retrieval balance $\alpha$ & $0.5$ \\
        Coarse retrieval budget $M$ & 20 \\
        MMR balance $\lambda$ & $0.7$ \\
        Activated skill number $|\mathcal{A}_{t}|$ & $5$ \\
        Backbone models & GPT-4.1, QWEN3-4B \\
        LLM-based components & Same as backbone model \\
        \bottomrule
    \end{tabular}
    }
    \label{tab:param_config}
\end{table}

\section{Additional Ablation Studies}
\label{app:additional_ablations}

We complement the factorial comparison in Section~\ref{sec:ablation} with ablations of the retrieval signal and skill extraction granularity.
The supplementary experiments in this appendix and Appendices~\ref{app:skill_library_analysis} and~\ref{app:inference_cost} use the same task sets and task order as the main evaluation and report single-run results.
Unless otherwise specified, they use \gpt on Shopping, Reddit, and Map.

\subsection{Retrieval Signals}
\label{app:ablation_retrieval}
All settings retain per-step retrieval and skill injection with MMR; task-only ($\alpha=1$) changes only the relevance signal. It differs from task-level static retrieval, which retrieves once without MMR and injects skills only into the initial context (Section~\ref{sec:ablation}).
We ablate the retrieval signal to study whether the task goal, the current webpage state, or their combination is most useful for selecting skills. As shown in \Cref{tab:ablation_retrieval}, task-only retrieval consistently outperforms state-only retrieval, suggesting that the initial task instruction remains an important anchor for skill selection. However, combining task and state information yields the best results across all three domains, with $\alpha=0.5$ achieving 34.6\%, 35.9\%, and 32.3\% on Shopping, Reddit, and Map, respectively. The lower performance at $\alpha=0.3$ and $\alpha=0.7$ further indicates that overemphasizing either the current state or the task goal is suboptimal.

\begin{table}[!ht]
    \centering
    \caption{Ablation study on retrieval signals with model \gpt on Shopping, Reddit, and Map websites. The best and second-best success rates within each domain are \textbf{bold} and \underline{underlined}, respectively.}
    \label{tab:ablation_retrieval}
    \resizebox{\columnwidth}{!}{%
    \begin{tabular}{lccc}
        \toprule
        Retrieval setting & Shopping & Reddit & Map \\
        \midrule
        Task-only ($\alpha=1$) & 30.3 & 32.6 & 30.8 \\
        State-only ($\alpha=0$) & 28.4 & 30.7 & 29.6 \\
        Task + state ($\alpha=0.3$) & 32.8 & \underline{34.5} & \underline{31.6} \\
        Task + state ($\alpha=0.5$) & \textbf{34.6} & \textbf{35.9} & \textbf{32.3} \\
        Task + state ($\alpha=0.7$) & \underline{33.4} & 34.1 & 31.2 \\
        \bottomrule
    \end{tabular}}
    \vspace{-0.7em}
\end{table}

\subsection{Skill Extraction Granularity}
\label{app:ablation_extraction}
We compare full-trajectory, single-action, and sliding-window skill extraction while keeping step-level task--state retrieval and MMR unchanged. The full-trajectory and sliding-window rows reproduce the dynamic conditions in \Cref{tab:rebuttal_factorial}; the single-action condition provides a finer-grained comparison. As shown in \Cref{tab:ablation_extraction}, sliding-window skills outperform both full-trajectory and single-action alternatives on all domains.
Full-trajectory skills preserve more task-level context but are less reusable for intermediate webpage states, leading to lower performance. Single-action skills perform worst because they
provide little abstraction over primitive browser actions to capture meaningful procedures.
In contrast, sliding-window extraction offers a better balance. It captures reusable multi-action sub-procedures while remaining flexible enough to be invoked at different execution states.

\begin{table}[!ht]
    \centering
    \scriptsize
    \caption{Ablation study on skill extraction granularity with backbone model \gpt on websites Shopping, Reddit, and Map. The best and second-best success rates within each domain are \textbf{bold} and \underline{underlined}, respectively.}
    \label{tab:ablation_extraction}
    \resizebox{\columnwidth}{!}{%
    \begin{tabular}{lccc}
        \toprule
        Extraction setting & Shopping & Reddit & Map \\
        \midrule
        Full Trajectory & \underline{31.6} & \underline{34.9} & \underline{31.4} \\
        Single Action & 29.5 & 24.7 & 25.4 \\
        Sliding Window & \textbf{34.6} & \textbf{35.9} & \textbf{32.3} \\
        \bottomrule
    \end{tabular}}
    \vspace{-0.8em}
\end{table}

\section{Skill Library Analysis}
\label{app:skill_library_analysis}

\subsection{Candidate Filtering and Skill Library Construction}
\label{app:candidate_filtering}
We analyze the filtering process by pooling all evaluator-judged successful trajectories from \gpt and \qwen on Shopping, Reddit, and Map.
\Cref{tab:candidate_filtering} reports the mean number of candidates per such trajectory at each stage; domains and models are pooled at the trajectory level rather than given equal weight.
On average, 25.4 enumerated windows yield 8.8 candidates passing LLM reusability judgment and 4.3 skills passing execution-based verification.
The final count is approximately 83.1\% lower than the raw candidate count.

\begin{table}[!ht]
    \centering
    \small
    \setlength{\tabcolsep}{3pt}
    \caption{Mean candidate counts per evaluator-judged successful trajectory, pooled across both backbones and Shopping, Reddit, and Map.}
    \label{tab:candidate_filtering}
    \begin{tabular}{lr}
        \toprule
        Stage & Candidates/trajectory \\
        \midrule
        Raw window enumeration & 25.4 \\
        After LLM reusability judgment & 8.8 \\
        After execution-based verification & 4.3 \\
        \bottomrule
    \end{tabular}
\end{table}

We enumerate valid starting positions for every window length in $\mathcal{L}=\{2,3,4,5\}$, allowing overlapping candidates to cover procedures that a single segmentation could split apart.
These lengths provide multi-action procedures while keeping them shorter than full-task trajectories.
Semantic validity is assessed by the LLM and subsequent trajectory-substitution verification described in \Cref{sec:method_skill_extraction}; window boundaries alone do not determine which candidates enter the library.
The single-action and full-trajectory comparisons in \Cref{tab:ablation_extraction} support this intermediate granularity.
Among retained skills, overlapping windows can still induce similar procedures, motivating MMR when selecting the skills exposed to the agent (\Cref{tab:ablation_mmr}).

\subsection{Skill Utilization}
\label{app:skill_utilization}
We measure how often the agent invokes retrieved skills during execution with \gpt across all five domains.
For task $i$, let $H_i$ be its number of agent decision steps and $K_i$ the number of those steps that invoke a retrieved skill.
We define utilization as
\begin{equation}
    \operatorname{Utilization}=\frac{1}{N}\sum_{i=1}^{N}\frac{K_i}{H_i}\times100\%,
\end{equation}
where $N$ includes all tasks in the domain, including failed tasks and early tasks for which the skill library is empty.
A skill invocation counts as one agent step regardless of how many primitive browser actions it executes.
The metric therefore averages the invocation fraction per task, rather than pooling steps across tasks.

\begin{table}[!ht]
    \centering
    \small
    \caption{Skill utilization with \gpt, averaged over all tasks within each domain.}
    \label{tab:skill_utilization}
    \begin{tabular}{lr}
        \toprule
        Domain & Utilization (\%) \\
        \midrule
        Shopping & 56 \\
        Admin & 47 \\
        Reddit & 38 \\
        GitLab & 31 \\
        Map & 53 \\
        \bottomrule
    \end{tabular}
\end{table}

As shown in \Cref{tab:skill_utilization}, the agent invokes skills in 31--56\% of its steps on a task-averaged basis, demonstrating active use of the learned procedures across all five domains.
Utilization is highest on Shopping and Map and lowest on GitLab.
The GitLab cases in Appendix~\ref{app:failure_cases} provide context for this lower adoption rate.

\section{Inference Cost and Runtime}
\label{app:inference_cost}
\Cref{tab:inference_cost} reports recorded LLM calls, token usage, and relative wall-clock time with \gpt, averaged over tasks pooled from Shopping, Reddit, and Map.
Token counts are reported as lower bounds, and runtime is normalized to ASI.
The additional state summarization and skill processing accompany higher call counts and runtime: \skillany records 9.4 calls per task and takes $1.57\times$ the runtime of ASI, illustrating the computational trade-off associated with its improved success rates.

\begin{table}[!ht]
    \centering
    \small
    \setlength{\tabcolsep}{4pt}
    \caption{Recorded computational cost with \gpt on Shopping, Reddit, and Map. Calls and tokens are averaged per task; runtime is relative to ASI.}
    \label{tab:inference_cost}
    \begin{tabular}{lrrr}
        \toprule
        Method & Calls/task & Tokens/task & Runtime \\
        \midrule
        Vanilla & 5.3 & $\geq19.7$k & $0.88\times$ \\
        AWM & 6.2 & $\geq20.2$k & $0.92\times$ \\
        ASI & 5.4 & $\geq21.9$k & $1.00\times$ \\
        CER & 6.8 & $\geq22.1$k & $1.28\times$ \\
        \skillany & 9.4 & $\geq23.3$k & $1.57\times$ \\
        \bottomrule
    \end{tabular}
\end{table}

\section{Case Studies}
\label{app:case_study}

We present representative skills induced by \skillany from five WebArena domains: Map, GitLab, Shopping, Reddit, and Admin.
Each skill is extracted from a judged-as-successful trajectory and represented as a parameterized code function paired with a natural-language description.
For example, the Map skill in Appendix~\ref{app:case_map_direction} fills the start and destination fields to submit a driving-directions query, while the GitLab skill in Appendix~\ref{app:case_gitlab_comment} fills and submits a merge-request comment.
Although these skills come from distinct websites, both separate webpage-specific element identifiers from task-specific content values, illustrating how \skillany learns reusable sub-procedures across different interaction patterns.
We first present the successful skill examples, then analyze representative failure cases on GitLab.

\subsection{Driving Directions Form Submission}
\label{app:case_map_direction}

The first skill is extracted from a Map task whose instruction is "Check if the social security administration in pittsburgh can be reached in one hour by car from CMU".
After the task is successfully completed, \skillany induces the following skill from the trajectory:
\begin{lstlisting}[style=skillstyle]
def submit_driving_directions_form(start_field_id, dest_field_id, go_button_id, start_location, destination):
    fill(start_field_id, start_location)
    fill(dest_field_id, destination)
    click(go_button_id)
\end{lstlisting}
The corresponding description is given as follows.
\begin{lstlisting}[style=promptstyle]
Fill in the starting point and destination fields and click the Go button to generate driving directions on a directions input form.
\end{lstlisting}
This skill is reusable because it separates structural webpage arguments, including \texttt{start\_field\_id}, \texttt{dest\_field\_id}, and \texttt{go\_button\_id}, from task-specific content arguments, namely \texttt{start\_location} and \texttt{destination}.
As a result, the same procedure can be invoked for future related map-navigation tasks when the current page satisfies the required conditions including input fields and submit button.

\subsection{Merge Request Comment Submission}
\label{app:case_gitlab_comment}

The second skill is extracted from a GitLab task whose instruction is to post ``lgtm'' for a merge request related to a specific project.
From this successful trajectory, \skillany induces the following skill:
\begin{lstlisting}[style=skillstyle]
def submit_merge_request_comment(comment_box_id, submit_button_id, comment):
    fill(comment_box_id, comment)
    click(submit_button_id)
\end{lstlisting}
Its description is:
\begin{lstlisting}[style=promptstyle]
Submit a comment on a merge request page by filling the comment textbox and clicking the submit button on a merge request detail view.
\end{lstlisting}
Although this skill comes from a different domain, it exhibits the same reusable abstraction pattern as the Map skill: element identifiers specify the current webpage structure, while the text argument specifies the task-dependent content.

Together, these examples show that \skillany can induce compact, parameterized skills that remain reusable across tasks while specifying the webpage conditions under which they apply.
They also illustrate the value of intermediate-state skill reuse: each procedure becomes useful when execution reaches a state where its required fields and buttons are available, even if that state occurs well after the task begins.

\subsection{Product Search and Wishlist Addition}
\label{app:case_shopping_wishlist}

The third skill is extracted from a Shopping task whose instruction is "Add Tide PODS Spring Meadow Scent HE Turbo Laundry Detergent Pacs, 81 Count to my wish list".
After the task is successfully completed, \skillany induces the following skill from the trajectory:
\begin{lstlisting}[style=skillstyle]
def search_and_add_first_product_to
    _wishlist(search_box_id, search_button_id,add_to_wishlist_button_id, product_query):
    fill(search_box_id, product_query)
    click(search_button_id)
    click(add_to_wishlist_button_id)
\end{lstlisting}
The corresponding description is given as follows.
\begin{lstlisting}[style=promptstyle]
Search for a product and add the first search result to the wish list on a product search results page.
\end{lstlisting}
This skill captures a longer e-commerce subroutine that combines product search, query submission, and wishlist addition.
It separates the task-specific content argument \texttt{product\_query} from structural webpage arguments, including \texttt{search\_box\_id}, \texttt{search\_button\_id}, and \texttt{add\_to\_wishlist\_button\_id}.
Compared with simpler two-step fill-and-submit skills, this example shows that \skillany can induce multi-step reusable procedures that abstract over repeated shopping interactions.

\subsection{Comment Reply Submission}
\label{app:case_reddit_reply}

The fourth skill is extracted from a Reddit task whose instruction is "Reply to the manager of the website in this post with 'thanks! I am a big fan of your website.'".
After the task is successfully completed, \skillany induces the following skill from the trajectory:
\begin{lstlisting}[style=skillstyle]
def submit_comment_reply(reply_box_id, post_button_id, message):
    fill(reply_box_id, message)
    click(post_button_id)
\end{lstlisting}
The corresponding description is given as follows.
\begin{lstlisting}[style=promptstyle]
Fill in a reply message and submit it using the reply textbox and post button on a comment thread page.
\end{lstlisting}
This skill represents a common social-forum interaction, where the agent fills a reply textbox and submits the response.
It separates the task-specific reply content \texttt{message} from structural webpage arguments, including \texttt{reply\_box\_id} and \texttt{post\_button\_id}.
Together with the GitLab merge-request comment skill, this example shows that similar fill-and-submit procedural patterns can emerge across different domains, such as forum discussion and code collaboration.

\subsection{Shipping Carrier Selection}
\label{app:case_admin_tracking}

The fifth skill is extracted from an Admin task whose instruction is "Update order \#306 with the UPS tracking number 55591023930".
After the task is successfully completed, \skillany induces the following skill from the trajectory:
\begin{lstlisting}[style=skillstyle]
def add_tracking_carrier(add_tracking_btn_id, carrier_dropdown_id, carrier_name):
    click(add_tracking_btn_id)
    select_option(carrier_dropdown_id, carrier_name)
\end{lstlisting}
The corresponding description is given as follows.
\begin{lstlisting}[style=promptstyle]
Select a shipping carrier from a dropdown after clicking the 'Add Tracking Number' button in the Shipping Information section on an order details page.
\end{lstlisting}
This skill captures an order-management operation in the Admin domain.
Unlike the previous examples that mainly rely on \texttt{fill} and \texttt{click}, this skill uses \texttt{select\_option} to choose a shipping carrier from a dropdown menu after expanding the tracking-number interface.
It separates the task-specific carrier argument \texttt{carrier\_name} from structural webpage arguments, including \texttt{add\_tracking\_btn\_id} and \texttt{carrier\_dropdown\_id}, showing that \skillany can induce reusable skills over different primitive action types.

Overall, these case studies show that \skillany learns compact procedural skills across all five WebArena domains.
The induced skills consistently separate webpage-specific structural arguments from task-specific content arguments, making them both grounded in the current page state and reusable for future tasks.
They also cover diverse interaction patterns, including form submission, comment posting, product search, wishlist addition, and dropdown selection.

\subsection{Failure Cases and GitLab Analysis}
\label{app:failure_cases}
We examine representative GitLab failures to understand how skill retrieval and execution interact with task requirements.

\paragraph{State--goal mismatch.}
In task 450, the agent must set the homepage URL on its GitLab profile.
Although the library contains a skill for setting the profile homepage URL, retrieval from the current project-listing dashboard favors project-browsing skills and fails to surface it.
This case illustrates a mismatch between skills applicable to the current page and procedures needed at a later task-relevant page: the profile-setting skill does not help the agent reach the page where it becomes applicable.

\paragraph{Irrelevant-skill injection.}
When the library lacks a suitable procedure, retrieval without a similarity threshold can still supply loosely related candidates.
Repository-creation tasks 752, 754, and 755, fork tasks 394 and 396, and member-invitation task 483 receive browsing skills with similarity scores of 0.33--0.44.
The agent typically ignores these candidates, but they introduce unnecessary options into the decision context.

\paragraph{Under-specified procedures.}
In task 306, the agent must count a user's commits between August and September 2022.
It invokes \texttt{filter\_commits\_by\_author}, which filters the author but does not enforce the date range.
The resulting reverse-chronological view leads the agent to continue scrolling until it exhausts its step budget.
A skill can therefore match part of the task while leaving a crucial constraint unsatisfied.

\paragraph{Skill coverage on GitLab.}
Persistent repository preconditions make operations such as forking, creating repositories from templates, and inviting members difficult, as discussed in \Cref{sec:main_results}.
Because skill induction depends on successful trajectories, unsuccessful operations also contribute fewer skills to subsequent tasks.
In the examined GitLab run, the six most frequently retrieved skills are all read/query procedures, while write operations are underrepresented.
This imbalance offers a qualitative explanation for the smaller gains on GitLab: the library provides more support for information-seeking tasks than for repository modifications.

\end{document}